\documentclass[letterpaper]{article} 
\usepackage{aaai2026}
\usepackage{amsmath}
\usepackage{mathtools}
\usepackage{amsfonts}
\usepackage{nicematrix}
\usepackage{multirow} 
\usepackage{float}
\usepackage{subfig}
\usepackage{booktabs}
\usepackage{times}  
\usepackage{helvet}  
\usepackage{courier}  
\usepackage[hyphens]{url}  
\usepackage{graphicx} 
\usepackage{natbib}  
\usepackage{caption} 
\usepackage{algorithm}
\usepackage{algorithmic}

\usepackage{newfloat}
\usepackage{listings}
\DeclareCaptionStyle{ruled}{labelfont=normalfont,labelsep=colon,strut=off} 
\floatstyle{ruled}
\newfloat{listing}{tb}{lst}{}
\floatname{listing}{Listing}
\title{A Tensor Residual Circuit Neural Network Factorized with Matrix Product Operation}
\author{
    Andi Chen
}
\affiliations{

}

\usepackage{bibentry}

\begin{document}

\maketitle
\begin{abstract}
\begin{quote}
It is challenging to reduce the complexity of neural networks while maintaining their generalization ability and robustness, especially for practical applications. 
Conventional solutions for this problem incorporate quantum-inspired neural networks with Kronecker products and hybrid tensor neural networks with MPO factorization and fully-connected layers. Nonetheless, the generalization power and robustness of the fully-connected layers are not as outstanding as circuit models in quantum computing. In this paper, we propose a novel tensor circuit neural network (TCNN) that takes advantage of the characteristics of tensor neural networks and residual circuit models to achieve generalization ability and robustness with low complexity.
The proposed activation operation and parallelism of the circuit in complex number field improves its non-linearity and efficiency for feature learning. Moreover, since the feature information exists in the parameters in both the real and imaginary parts in TCNN, an information fusion layer is proposed for merging features stored in those parameters to enhance the generalization capability. Experimental results confirm that TCNN showcases more outstanding generalization and robustness with its average accuracies on various datasets 2\%-3\% higher than those of the state-of-the-art compared models. More significantly, while other models fail to learn features under noise parameter attacking, TCNN still showcases prominent learning capability owing to its ability to prevent gradient explosion. Furthermore, it is comparable to the compared models on the number of trainable parameters and the CPU running time. An ablation study also indicates the advantage of the activation operation, the parallelism architecture and the information fusion layer.
\end{quote}
\end{abstract}

\section{Introduction}
With the rapid development of neural networks and tensor networks, tensor neural network, a neural network designed to process tensor data directly, leverages the multi-dimensional structures of tensors to capture complex features \cite{5,9,27,33,69,68,67,66*}. By matrix factorization and decomposition, it usually possesses fewer trainable parameters and consume fewer resources than classical neural networks with large-sized hidden layers \cite{34,35,36,37,67,68}. Nevertheless, one problem lies in its limited generalization power and robustness due to its small parameter manifold spaces, which implies that their capability of pattern learning and interference resistance requires to be further facilitated \cite{27}. Meanwhile, in quantum computing, residual circuit model demonstrates outstanding fault tolerance, high fidelity and strong generalization and robustness for feature extraction based on their logical unitary gates on the quantum-inspired deep learning field \cite{43,44,45,46,47,70,71,80,81}. Nevertheless, owing to continuous multiplications of sparse matrices, the time complexity of circuit models, usually proportional to the exponential power of the neural network width, is relatively higher than that of classical neural networks \cite{2,46,86,81}.

Hence, to improve the generalization power, robustness while reducing complexity of feedforward neural networks in the meantime, in this paper, we propose a \textbf{T}ensor \textbf{C}ircuit \textbf{N}eural \textbf{N}etwork (TCNN) with matrix product operators and residually continuous circuit architecture. Clean, enhanced and noisy MNIST, FASHIONMNIST, CIFAR, ImageNet and CelebA datasets are for the assessment of generalization capability and robustness of TCNN \cite{50,51,61,88,89}. Besides that, we also select another several state-of-the-art feedforward neural networks incorporating a \textbf{M}ulti-\textbf{L}ayer \textbf{P}erceptron (MLP), a \textbf{H}ybrid \textbf{F}eedforward \textbf{T}ensor \textbf{N}eural \textbf{N}etwork (HFTNN) \cite{27}, a \textbf{H}ybrid \textbf{F}eedforward \textbf{C}ircuit \textbf{N}eural \textbf{N}etwork (HFCNN) \cite{46}, a \textbf{Q}uantum-\textbf{i}nspired \textbf{M}ulti-\textbf{L}ayer \textbf{P}erceptron (QiMLP) \cite{75*}, a \textbf{F}actor \textbf{A}ugmented \textbf{T}ensor-on-\textbf{T}ensor \textbf{N}eural \textbf{N}etwork (FATTNN)\cite{66*}, a \textbf{H}ybrid \textbf{Q}uantum-classical convolutional \textbf{Net}work (HQNet)\cite{90}, a residual and a dense convolutional neural networks with spatial and channel attentions (Resnet and Densenet) \cite{68} for model comparison. Experiments demonstrate that TCNN possess benefits on complexity, generalization capability, robustness and stability. Hence, the contributions in this paper are as follows:
\begin{itemize}
\item We devise a tensor circuit neural network that is comprised of a tensor network with MPO factorization with paralleled adaptive residual circuit framework in complex number field;
\item The parameter complexity of TCNN is lower than or comparable to that of the compared models;
\item Experiments showcase that our TCNN is more advantageous than all the compared models on generalization power and accuracy on image recognition. 
\item Experiments indicate that our TCNN showcases much more outstanding robustness to random noise parameter attacks than all the compared models. The reason lies in the strong capability of it to prevent gradient explosion. 
\end{itemize}

\section{Related work}
\textbf{Tensor neural networks.} In tensor neural networks, through various decomposition methods, high-dimensional matrices could be decomposed into low-dimensional parameterized tensors that possess learning capability and are stored in lists or tuples without taking up too much resource. Common methods incorporate Block-Term Tucker (BTT) decomposition, Matrix Product State (MPS) decomposition, Tensor Train (TT) decomposition, Matrix Product Operator (MPO) factorization, etc. while other methods need to solve the eigenvalues or singular values, MPO factorization, allowing parallel computation of local tensor operations, is more efficient without computing those values \cite{38,65,66*,67,68,91}. Compared with Kronecker method, it saves more time with strong explainability \cite{39,66*}. Hence, impacted by MPO method for compressing deep neural networks \cite{39}, we utilize an MPO factorization by constituting several small-sized activated parameter matrices for feature learning, which may also reduce the number of trainable parameters and save training and testing time. \\
\textbf{Residual circuit models.}
\begin{figure}[!t]
	\centering
	\includegraphics[width=3.2in]{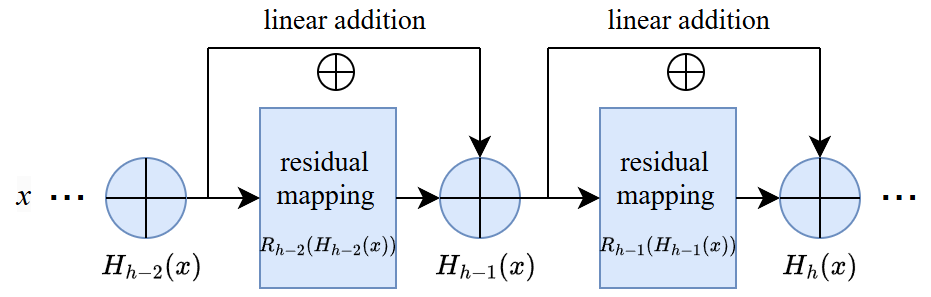}
	\caption{Residual connections. As for the proposed residual circuit model in this paper, the residual mapping is continuous unitary transformation with activations.}
\end{figure}
As is shown on fig.1, through direct linear addition between adjacent layers $H_{h-1}(x)$ and $H_h(x)$, the residual network effectively transfers the main features in the data forward, thereby improving the accuracy and generalization ability. Hence, some works have explored the performance of residual circuit models that are capable of both capturing data patterns and preventing gradient explosion, which improves their ability of resisting noise interferences \cite{46,47,48,49,55}. However, in these works, the circuit model only represents linear mapping and possess serial architectures. And most of them don't involve complex number. Simultaneously, a few works demonstrate that paralleled circuit networks and complex neural networks are able to capture patterns with different details, thus improving generalization power \cite{2,71,72,73,80}. Therefore, we propose parallel non-linear circuit models with activations in complex number field, which may further assist to capture important data features develop neural networks in complex number field.
\begin{figure*}[!t]
	\centering
	\includegraphics[width=7in]{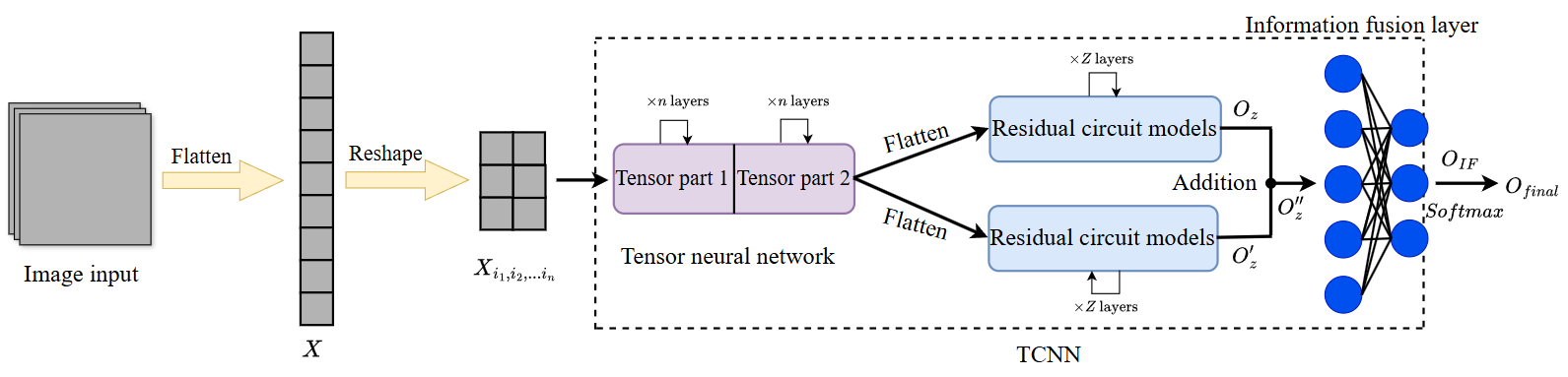}
	\caption{Architecture of TCNN and the process of transferring data into TCNN. TCNN incorporates one tensor neural network architecture with two tensor parts and paralleled circuit framework.}
\end{figure*}
\begin{table*}[!t]
	\centering
	\renewcommand{\arraystretch}{0.4}
	\caption{Number of trainable parameters in all the models}
	\label{table_example}
	\centering
	\scalebox{0.8}{
	\begin{tabular}{c c c c c c c}
		\toprule
		  MLP & QiMLP & FATTNN & HQNet & HFTNN & HFCNN & TCNN \\
		\midrule
		$2.67\times10^5$ & $1.72\times10^3$ & $5.58\times 10^3$& $2.36\times10^3$ & $4.11\times10^3$ & $2.02\times10^5$ & $4.09\times10^3$ \\
		\bottomrule
	\end{tabular}}
\end{table*}
\section{Preliminaries}
\textbf{Matrix product operator factorization.} Suppose $N_x$ and $N_y$ are respectively the dimensions of $X$ and $Y$. To construct the factorized matrices of the MPO method, we firstly reshape $W$ into a 2$n$-indexed tensor:
\begin{equation}
    W = W_{j_1j_2...j_n,i_1i_2...i_n}.
\end{equation}
Here, the one-dimensional input $X$ with dimension $N_x$ is reshaped into a tensor of $n$ order, $X_{i_1,i_2,...,i_n}$, with the index of the $k^{th}$ order being $i_k (k=1,2,...n)$. In the meantime, the one-dimensional output $Y$ with dimension $N_y$ is reshaped into a tensor of $n$ order $Y_{j_1,j_2,...,j_n}$ as well, with the index of the $k^{th}$ order being $j_k$($k=1,2,...n$). Suppose $I_k$ and $J_k$ are the dimensions of $i_k$ and $j_k$ separately, we have:
\begin{equation}
    \prod \limits_{k=1}^n I_k = N_x, \\\\\ \prod \limits_{k=1}^n J_k = N_y.
\end{equation}
Besides that, the total number of trainable parameters in the MPO factorization process, $P_{MPO}$, is:
\begin{equation}
    P_{MPO} = \prod^{n-1}_{k=2}(I_kJ_kD^2) + (I_1J_1+I_nJ_n)D + N_y
\end{equation}
\textbf{Circuit evolution.} In our framework, the $k^{th}$ evolution process stands for the manipulation to the vector with continuous unitary operations \cite{46}:
\begin{equation}
	\label{deqn_ex1a}
	O_{k+1} = U_{k} O_{k} \quad(k \in \mathbb{N}^+),
\end{equation}
where $U_k$ represents the $k^{th}$ unitary gate in the $k^{th}$ evolution and $O_{k+1}$ denotes the output vector of the $k^{th}$ evolution. The gate $T_{m}(\phi_r^l,\theta^{l}_r)$, frequently utilized in $U_k$, can be written in the computational basis as the following matrix
\begin{equation}
	\begin{tiny}
		\label{deqn_ex1a}	
		\begin{pNiceMatrix}[first-col,last-row]
			&   1  & 0    &\cdots &\cdots & \cdots & \cdots &\cdots & 0    \\
			&   0  & 1    &      &           &            &      &   &  \vdots    \\
			& \vdots &      &\ddots&           &            &      &   & \vdots     \\
			m   &\vdots&      &      &e^{i\phi^l_r}\cos({\mathrm\theta}^{l}_r)&{\rm -sin}(\theta^{l}_r)&      &   & \vdots     \\
			m+1  &\vdots	    &      &      &e^{i\phi^l_r}\sin({\rm\theta}^{l}_r)&\cos({\rm\theta}^{l}_r) &    &   &\vdots\\
			&\vdots	    &      &      &           &            &\ddots&   & \vdots     \\
			&\vdots	    &      &      &   & &      & 1 & 0    \\
			&    0&\cdots&\cdots &\cdots&\cdots &\cdots& 0 & 1    \\
			&     &      &      &   m       &  m+1       &      &   &      \\
		\end{pNiceMatrix}
		.
	\end{tiny}
\end{equation}
As for this matrix, $m$ denotes the row index of the term ${\rm -sin}({\theta}^{l}_r)$, also the column index of the term $\sin({\theta}^{l}_r)$ $(m \in \mathbb{N}^+)$. $\theta^{l}_r$ and $\phi_r^l$ denote the $(r+1)^{th}$ parameter of the real part and the imaginary part separately in the $l^{th}$ circuit layer $(r \in \mathbb{N}, l \in \mathbb{N}^+)$. $i$ represents the imaginary unit. And the relation between $r$ and $m$ is given by:
\begin{equation}
	r=
	\begin{cases}
		m-1& \text{ $ 0 \leq r \leq N-2 $ } \\
		2N-3-m& \text{ $ N-2 < r \leq 2N-4, $ }
	\end{cases}
\end{equation}
where $N$ is the dimension of the circuit layer.
\section{Methodology}
In this section, we illustrate our TCNN which incorporate details of tensor layers and residual circuit layers. Supplementary material shows the frameworks of tensor neural network on the basis of MPO factorization and one layer of the circuit model with unitary evolution and activation. Fig.2 showcases the framework of TCNN comprised of a tensor neural network framework and paralleled residual circuit model. \\
\textbf{Framework of TCNN.} To begin with, the input image data is firstly flattened into a vector $X$, which is then transferred into the first tensor part with $n$ tensor layers. Additionally, to extract its feature, we reshape $X$ into a $n$-order tensor $X_{i_1,i_2,...,i_n}$. We also set the dimension of the expected output $Y$ and reshape it into another $n$-order tensor $Y_{j_1,j_2,...,j_n}$. The MPO representation of weight matrix $W$ is gained by factorizing it into $n$ local tensors that are stored in a list as well:
\begin{equation}
    W_{j_1j_2...j_n,i_1i_2...i_n} = list(w^{(1)}[j_1,i_1],...,w^{(n)}[j_n,i_n]),
\end{equation}
where $w^{(k)}[j_k,i_k]$ is small-sized parameterized local matrices. We also define the MPO bond vector $D_{bond} = (D_0, D_1,...,D_{n})$ to link the local tensors in $W$ and we have $w^{(k)}[j_k,i_k] \in \mathbb{R}^{J_kD_k\times I_kD_{k-1}}$ \cite{39}. For computation convenience, we suppose $D_0 = D_n = 1$ and $D_k = D(k = 1,...,n-1)$, then $w^{(1)}[j_1,i_1] \in \mathbb{R}^{J_1D \times I_1}, w^{(k)}[j_k,i_k] \in \mathbb{R}^{J_kD \times I_kD}(k \neq 0, k \neq n), w^{(n)}[j_n,i_n] \in \mathbb{R}^{J_n \times I_nD}$. In the MPO computing process, $X_{i_1,i_2,...,i_n}$ is reshaped into matrices with various sizes many times based on $I_k$. Suppose $X_{i_1,i_2,...,i_n} = (X^{(1)}, X^{(2)},..., X^{(n)})$, where $X^{(k)}$ is the matrix obtained by reconstructing $X_{i_1,i_2,...,i_n}$ for the $k^{th}$ time. And $X^{(1)} \in \mathbb{R}^{I_1\times I_2I_3...I_n}, X^{(k)} \in \mathbb{R}^{I_kD\times J_1...J_{k-1}I_{k+1}...I_n}$. Suppose the output of the $k^{th}$ tensor layer is $Y^{k}$, we have:
\begin{equation}
    Y^{k} = w^{(k)}[j_k,i_k] X^{(k)} (k=1,2,...,n).
\end{equation}
$Y^{k} \in \mathbb{R}^{J_kD\times J_1..J_{k-1}I_{k+1}...I_n} (k \neq n), Y^{n} \in \mathbb{R}^{J_n\times J_1..J_{n-1}}$. And we reshape $Y^{n}$ to be a vector $Y^{n}_{final} \in \mathbb{R}^{J_1J_2...J_n}$, eventually we get:
\begin{equation}
    Y^{n}_{final} = \sigma(Y^{n}+b),
\end{equation}
where $b$ represents bias. To further reduce the dimension of the output of the first tensor part, we transfer it into the second tensor part with $n$ tensor layers and local weight matrices as well. The final output vector of the second tensor part is transmitted into the first layer of the paralleled circuit models, which is abbreviated as $O_1$ now. 
\begin{figure}[!t]
	\hspace{-0.25cm}
    \includegraphics[width=3.45in]{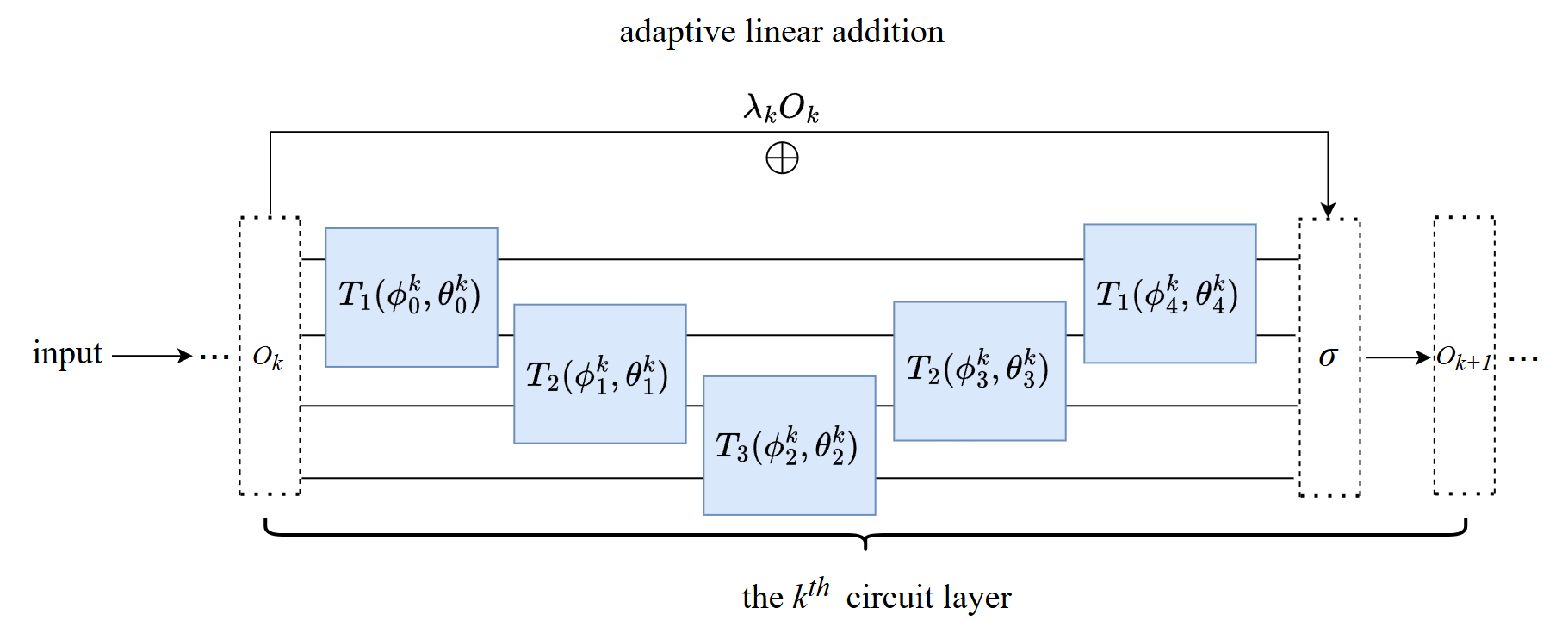}
	\caption{One layer of the residual circuit models with activations.$O_k$ and $O_{k+1}$ separately denote the input and output of the $k^{th}$ circuit layer. $\sigma$ represents any activation function. $\phi^k_r$ and $\theta^k_r$ represent trainable parameters in the real and imaginary parts. They are updated by back propagation method.}
\end{figure}
For the circuit model, fig.3 demonstrates the $k^{th}$ circuit layer with its dimension being 4. The “V" shape circuit assists feature learning avoiding learning imbalance \cite{46}. The adaptive parameters can effectively control the impact of noise in the data on feature learning \cite{56}. We define:
\begin{equation}
	\begin{small}
		U_{k} \coloneqq (\prod \limits_{\substack{j=0}}^{\substack{j=N-2}}T_{j+1}(\phi_{j}^{k},\theta_{j}^{k})) \times (\prod \limits_{\substack{j=N-1}}^{\substack{j=2N-4}}T_{2N-3-j}(\phi_{j}^{k},\theta_{j}^{k})),
    \end{small}
\end{equation}
where $U_k$ is the product of the continuous multiplications of $T$ matrices in the $k^{th}$ circuit layer. Suppose there is $Z$ circuit layers with $N$ dimensions in one circuit model. Then for the output vector of the $k^{th}$ circuit layer of the two paralleled circuits, $O_{k+1}$ and $O_{k+1}^{\prime}$, we have:
\begin{equation}
\begin{aligned}
	O_{k+1} = \sigma(U_{k}O_k \oplus \lambda_{k}O_k) \\
    O_{k+1}^{\prime} = \sigma(U_{k}^{\prime}O_k^{\prime} \oplus \lambda_{k}^{\prime}O_k^{\prime})
\end{aligned}
\end{equation}
where $\oplus$ denotes adaptive residual addition. The formula of $U_{k}^{\prime}$ is identical to that of $U_{k}$. $\lambda_{k}$ and $\lambda_{k}^{\prime}$ are adaptive parameters in the two paralleled circuits. To leverage the upside of the parallelism of circuit models, we transfer two identical $O_1$ into the two identical circuit models, gaining the output complex vectors of the two circuit models respectively, $O_Z$ and $O_Z^{\prime}$. In $O_Z$, the vectors consisting of the values of the real part of all elements and the values of the imaginary part of all elements are denoted as $O_{Z_1}$ and $O_{Z_2}$ respectively. In $O_Z^{\prime}$, the vector consisting of the values of the real part of all elements and the values of the imaginary part of all elements is denoted as $O_{Z_1}^{\prime}$ and $O_{Z_2}^{\prime}$ separately. As fig.4 shows, We merge the information of $O_{Z_1}$ and $O_{Z_1}^{\prime}$, $O_{Z_2}$ and $O_{Z_2}^{\prime}$ by linear addition, that is:
\begin{equation}
\begin{aligned}
    &O_Z = O_{Z_1} + O_{Z_2}i\\
    &O_Z^{\prime} = O_{Z_1}^{\prime} + O_{Z_2}^{\prime}i\\
    &O_{Z_1}^{\prime \prime} = O_{Z_1}+O_{Z_1}^{\prime} \\
    &O_{Z_2}^{\prime \prime} = O_{Z_2}+O_{Z_2}^{\prime} \\
    &O_{Z}^{\prime \prime} = concate(O_{Z_1}^{\prime \prime}, O_{Z_2}^{\prime \prime}),
\end{aligned}
\end{equation}
\begin{figure}[!t]
	\hspace{-0.3cm}
    \includegraphics[width=3.45in]{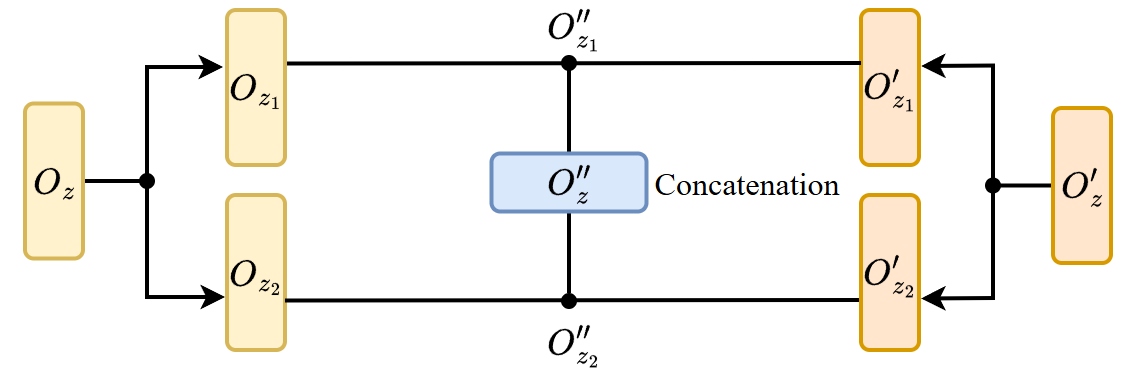}
	\caption{Process of merging information of the real and imaginary parts of the paralleled circuit in the information fusion layer.}
\end{figure}
where $concate$ represents concatenation of the two vectors on the column index. $O_{Z_1}^{\prime \prime}$ and $O_{Z_2}^{\prime \prime}$ store different features extracted by the real and imaginary parts of the circuit models respectively. $i$ is the complex number unit. Then to capture the important patterns in the two vectors, they are concatenated before transferred into the information fusion layer with fully-connected frameworks. We have:
\begin{equation}
    \begin{aligned}
        &O_{IF} = \sigma(W_{IF} \cdot O_z^{\prime \prime}+B_{IF}) \\
        &O_{final} = softmax(W_{out} \cdot O_{IF}+B_{out}),
    \end{aligned}
\end{equation}
where $W_{IF}$ and $B_{IF}$ are weights and biases in the information fusion layer, $W_{out}$ and $B_{out}$ are weights and biases in the output layer. $softmax$ is the softmax activation function of the output layer \cite{2}. $O_{IF}$ and $O_{final}$ are the output of the information fusion layer and the output layer separately. \\
\textbf{Gradient descent method in complex number field.}
Some studies have demonstrated gradient descent method in complex number field with various methods such as enhanced gradient descent with Hessian matrices, measurement gradient descent with observables and differential gradient descent \cite{62,63,64}. Contemplating the complexity and convenience, we utilize stochastic gradient descent (SGD) method with momentum to update parameters in the real and imaginary parts in our TCNN. Suppose $loss$ denotes the loss function of our TCNN, for $\theta^k_r$ and $\phi^k_r$ in the $k^{th}$ layer of which the output is $O_{k+1}$, we have:
\begin{equation}
    \begin{aligned}
        &v_t = m_c v_{t-1} + (1-m_c)\frac{\partial (loss)}{\partial \theta^k_r}\\
        &v_t^{\prime} = m_c v_{t-1}^{\prime} + (1-m_c)\frac{\partial (loss)}{\partial \phi^k_r}\\
        & \theta^k_r \leftarrow \theta^k_r - \eta v_t \\
        & \phi^k_r \leftarrow \phi^k_r - \eta v_t^{\prime},
    \end{aligned}
\end{equation}
where $v_t$ and $v_t^{\prime}$ are the momentum factors for the $t^{th}$ update of $\theta^k_r$ and $\phi^k_r$ separately. $m_c$ and $\eta$ are the momentum coefficient and learning rate respectively. from Eq.(11), we could compute $\frac{\partial O_{(k+1)}}{\partial U_k}$, $\frac{\partial O_{(k+1)}}{\partial O_{k}}$. From the chain rule \cite{2}, we have:
\begin{equation}
    \begin{aligned}
        & \frac{\partial (loss)}{\partial O_{Z}} = \frac{\partial (loss)}{\partial O_{final}} \cdot \frac{\partial O_{final}}{\partial O_{IF}} \cdot \frac{\partial O_{IF}}{\partial O_{Z}^{\prime \prime}} \cdot \frac{\partial O_{Z}^{\prime \prime}}{\partial O_{Z}} 
    \end{aligned}
\end{equation}
According to Eq.(6), there are two occasions of the derivative of $O_{k+1}$ to the parameters. When $0 \leq r \leq N-2$, we also have:
\begin{subequations}
    \begin{small}
	\begin{align}
		&\frac{\partial U_{k}}{\partial \theta_r^{k}}  = (\prod \limits_{j=0}^{r-1}T_{j+1}(\phi_j^k,\theta_j^{k})) \cdot G_{r+1}(\theta_r^{k}) \cdot (\prod \limits_{j=r+1}^{N-2}T_{j+1}(\phi_j^k,\theta_j^{k})) \nonumber\\
		&\cdot (\prod \limits_{j=N-1}^{2N-4}T_{2N-3-j}(\phi_j^k,\theta_j^{k}))\\
		&\frac{\partial U_{k}}{\partial \phi_r^{k}} = (\prod \limits_{j=0}^{r-1}T_{j+1}(\phi_j^k,\theta_j^{k})) \cdot G_{r+1}(\phi_r^{k}) \cdot (\prod \limits_{j=r+1}^{N-2}T_{j+1}(\phi_j^k,\theta_j^{k})) \nonumber\\
		&\cdot (\prod \limits_{j=N-1}^{2N-4}T_{2N-3-j}(\phi_j^k,\theta_j^{k})), 
	\end{align}
    \end{small}
\end{subequations}
where $G_{r+1}(\theta_r^{k})$ and $G_{r+1}(\phi_r^{k})$ are the first derivative matrices of $\theta_r^{k}$ and $\phi_r^{k}$ respectively. And when $N-2 < r \leq 2N-4$, we have:
\begin{subequations}
	\begin{small}
    \begin{align}
		&\frac{\partial U_{k}}{\partial \theta_r^{k}}= (\prod \limits_{j=0}^{N-2}T_{j+1}(\phi_j^{k},\theta_j^{k})) \cdot (\prod \limits_{j=N-1}^{r-1}T_{2N-3-j}(\phi_j^k,\theta_j^{k})) \nonumber\\
		&\cdot G_{2N-r-3}(\theta_r^{k}) \cdot (\prod \limits_{j=r+1}^{2N-4}T_{2N-3-j}(\phi_{j}^k,\theta_j^{k})) \\
		&\frac{\partial U_{k}}{\partial \phi_r^{k}}= (\prod \limits_{j=0}^{N-2}T_{j+1}(\phi_j^{k},\theta_j^{k})) \cdot (\prod \limits_{j=N-1}^{r-1}T_{2N-3-j}(\phi_j^k,\theta_j^{k})) \nonumber\\
		&\cdot G_{2N-r-3}(\phi_r^{k}) \cdot   (\prod \limits_{j=r+1}^{2N-4}T_{2N-3-j}(\phi_{j}^k,\theta_j^{k})) 
    \end{align}
	\end{small}
\end{subequations}
Therefore, the derivative of the loss function to the parameters can be written as:
\begin{subequations}
    \begin{align}
        & \frac{\partial (loss)}{\partial \theta^k_r} = \frac{\partial (loss)}{\partial O_{Z}} \cdot (\prod_{j=Z}^{k+2}\frac{\partial O_{j}}{\partial O_{j-1}}) \cdot \frac{\partial O_{k+1}}{\partial U_k} \cdot \frac{\partial U_k}{\partial \theta^k_r} \\
        & \frac{\partial (loss)}{\partial \phi^k_r} = \frac{\partial (loss)}{\partial O_{Z}} \cdot (\prod_{j=Z}^{k+2}\frac{\partial O_{j}}{\partial O_{j-1}}) \cdot \frac{\partial O_{k+1}}{\partial U_k} \cdot \frac{\partial U_k}{\partial \phi^k_r},
    \end{align}
\end{subequations}
According to previous work \cite{46}, most of the terms in Eq.(18a) and (18b) involve plenty of cosine and sine functions. Due to their limited value range, the derivatives are not very large, thus preventing gradient explosion systematically, while other neural networks don't possess this capability. Besides that, a common problem in circuit model is gradient disappearance \cite{81}. Nevertheless, since residual connection could prevent zero gradient \cite{2}, it is relatively easy to train the forward and back propagation of our TCNN. What's more, the update law of parameters in the tensor neural network and another circuit model is similar with that in the circuit model example. 
\\
\textbf{Parameter complexity.} Parameter complexity incorporates the parameters that are temporarily stored in each epoch and stored throughout the training and testing processes. According to previous work (\cite{3,46}), we deduce the parameter complexity of the circuit model $P_{CM}$ separately contemplating the parameters that could store key information of data features. We have:
\begin{equation}
    \begin{aligned}
        P_{CM}&= O(2Z\cdot (2N-3))+O(2Z\cdot (2N-4)) \\
			&+O(Z-1)+O(Z-2) \\
        &\approx  O(8NZ) \\ 
        &\approx O(NZ),
    \end{aligned}
\end{equation}
where $O(2Z\cdot (2N-3))$ denotes the number of parameters in the real and imaginary parts in the forward propagation process. $O(2Z\cdot (2N-4))$ denotes the number of the gradients of the parameters in the real and imaginary parts in the back propagation process. $O(Z-1)$ and $O(Z-2)$ are the number of adaptive parameters and the number of their gradients respectively. As for the fully-connected information fusion layer, suppose the matrix size of it is $P_{IF_1} \times P_{IF_2}$, the parameter complexity of the layer, $P_{IF}$, is:
\begin{equation}
        P_{IF}= P_{IF_1}\cdot P_{IF_2}
\end{equation}
For TCNN, the parameter complexity, $P_{TCNN}$ is:
\begin{equation}
        P_{TCNN}= P_{MPO}+2P_{CM}+P_{IF}
\end{equation}
 The formula of parameter complexity of our TCNN is similar with that of the models in the previous work \cite{46}. Associated with Eq.(3), the parameter complexity of the compared models is given in table 1. Compared with the MLP, FATTNN and HFCNN, we reduce the number of trainable parameters with MPO method and circuit models in TCNN to a certain extent. In the following experiments, we need to verify whether our TCNN possesses strong generalization ability and robustness without too many parameters.

\section{Experiments}
We conduct a few groups of experiments with clean, enhanced and noisy datasets. The indicators include accuracy, loss and CPU running time \cite{57}. Details of the baselines are in the supplementary material.\\
\textbf{Experiment setup.} 
We use Pytorch, single runs and SGD optimizer to accomplish our experiments. As for the enhanced datasets, we set their brightness, contrast, saturation and hue all being random numbers close to 0.5. For the noisy datasets, we add standard normal distribution or uniform noise, which usually occurs in the engineering field \cite{58}, into the images to evaluate the robustness of our proposed network. All the results are testing results.\\
\textbf{Generalization power test.}
\begin{table*}[!t]
	\centering
	\renewcommand{\arraystretch}{0.75}
	\caption{Average accuracy,loss values and CPU running time of the seven models with clean and enhanced datasets. * means enhanced datasets.}
	\label{table_example}
	\centering
	\scalebox{0.75}{
    \setlength{\tabcolsep}{1mm}
	\begin{tabular}{ccccccccc}
		\toprule
		&Dataset & MLP & QiMLP & FATTNN& HQNet & HFTNN & HFCNN & TCNN \\
		\midrule
		&MNIST & 96.62\%$\pm$0.61\% &98.14\%$\pm$0.41\%&97.35\%$\pm$0.49\%&95.85\%$\pm$0.68\%& 96.88\%$\pm$0.60\% & 97.31\%$\pm$0.46\% & \textbf{98.64\%$\pm$0.39\%} \\
		\multirow{8}{*}{Accuracy}&FASHIONMNIST & 92.24\%$\pm$0.90\% &93.64\%$\pm$0.79\%&92.06\%$\pm$0.92\%&91.12\%$\pm$0.95\%& 91.67\%$\pm$0.93\% & 92.51\%$\pm$0.88\% & \textbf{93.89\%$\pm$0.76\%} \\
		&MNIST*      & 97.86\%$\pm$0.44\% &\textbf{99.26\%$\pm$0.37\%}&97.56\%$\pm$0.47\%&96.59\%$\pm$0.61\%& 99.01\%$\pm$0.37\% & 97.95\%$\pm$0.43\% & 99.03\%$\pm$0.40\% \\
        &FASHIONMNIST* & 92.08\%$\pm$0.91\% &93.26\%$\pm$0.74\%&92.71\%$\pm$0.83\%&91.79\%$\pm$0.94\%& 91.76\%$\pm$0.92\% & 91.89\%$\pm$0.94\% & \textbf{93.64\%$\pm$0.72\%} \\
        &ImageNet & 76.89\%$\pm$0.98\% &80.03\%$\pm$0.77\%&78.54\%$\pm$0.87\%&77.54\%$\pm$0.91\%& 76.88\%$\pm$0.98\% & 78.90\%$\pm$0.86\% & \textbf{81.44\%$\pm$0.72\%} \\
		&CelebA & 91.43\%$\pm$0.99\% &92.01\%$\pm$0.92\%&92.00\%$\pm$0.92\%&92.95\%$\pm$0.85\%& 90.88\%$\pm$0.93\% & 91.91\%$\pm$0.97\% & \textbf{93.04\%$\pm$0.72\%} \\
		&ImageNet* & 76.45\%$\pm$0.99\% &79.38\%$\pm$0.81\%&78.08\%$\pm$0.88\%&77.99\%$\pm$0.90\%& \textbf{80.45\%$\pm$0.78\%} & 77.12\%$\pm$0.96\% & 80.31\%$\pm$0.83\% \\
        &CelebA* & 92.06\%$\pm$0.95\% &93.16\%$\pm$0.72\%&92.18\%$\pm$0.90\%&92.55\%$\pm$0.88\%& 91.56\%$\pm$1.01\% & 91.11\%$\pm$1.08\% & \textbf{93.45\%$\pm$0.71\%} \\
        \midrule
        &MNIST & 0.0038 &0.0026&0.0031&0.0039& \textbf{0.0021} & 0.0024 & 0.0022 \\
		\multirow{8}{*}{Loss}&FASHIONMNIST & 0.0076 &0.0064&0.0069&0.0078& 0.0065 & 0.0068 & \textbf{0.0062} \\
		&MNIST*      & 0.0036 &\textbf{0.0024}&0.0028&0.0039& 0.0029 & 0.0026& \textbf{0.0024} \\
        &FASHIONMNIST* & 0.0069 &0.0071&0.0073&0.0078& 0.0073 & 0.0070 & \textbf{0.0068} \\
        &ImageNet & 0.0849 &0.0826&0.0815&0.0857&0.0896 & 0.0740 & \textbf{0.0727} \\
		&CelebA & 0.0134 &\textbf{0.0103}&0.0199&0.0173& 0.0134 & 0.0188 & 0.0141 \\
		&ImageNet*  & 0.0889 &0.0875&0.0846&0.0887&0.0837 & 0.0804 & \textbf{0.0802} \\
        &CelebA* & 0.0189 &0.0157&0.0195&0.0157& 0.0166 & 0.0177 & \textbf{0.0150} \\
        \midrule
        \multirow{4}{*}{CPU running time}&MNIST & 612s &\textbf{394s}&513s&653s& 426s & 853s & 634s \\
		&FASHIONMNIST & 692s &\textbf{476s}&612s&719S& 512s & 802s & 689s \\
		&MNIST*  & 698s &\textbf{489s}&721s&756s& 504s & 946s& 745s \\
        &FASHIONMNIST* & 685s &\textbf{430s}&687s&734s& 470s & 894s & 714s \\
		\bottomrule
	\end{tabular}}
\end{table*}
\begin{table*}[!t]
	\centering
	\renewcommand{\arraystretch}{0.8}
	\caption{Variances of the seven models with enhanced and noisy datasets}
	\centering
	\scalebox{0.8}{
    \setlength{\tabcolsep}{7mm}
	\begin{tabular}{c c c c c c c c}
		\toprule
		Dataset & MLP & QiMLP & FATTNN& HQNet & HFTNN & HFCNN & TCNN \\
		\midrule
		MNIST& 0.0652 & 0.0378&0.0390&0.0829&0.0482 & 0.0977 & \textbf{0.0249} \\
		FASHIONMNIST& 0.0383 & 0.0289&0.0531&0.0712& 0.0515 & 0.0417 & \textbf{0.0204} \\
		CIFAR10& 0.1076 &0.0631&0.0865&0.0942& 0.1086 & 0.0689 & \textbf{0.0582} \\
        CIFAR100& 0.0946 &0.0478&0.0745&0.0834& 0.0825 & 0.0792 & \textbf{0.0429} \\
        ImageNet& 0.1653 &0.1197&0.1245&0.1608& 0.1466 & 0.1551 & \textbf{0.1164} \\
        CelebA& 0.0675 &0.0419&0.0501&0.0665& 0.0496 & 0.0592 & \textbf{0.0405} \\
		\bottomrule
	\end{tabular}}
\end{table*}
\begin{table*}
	\centering
	\renewcommand{\arraystretch}{0.6}
	\caption{Accuracy of the seven models to random normal distribution noise attacks}
	\label{table_example}
	\centering
	\scalebox{0.8}{
    \setlength{\tabcolsep}{1mm}
	\begin{tabular}{c c c c c c c c}
		\toprule
		Dataset & MLP & QiMLP & FATTNN& HQNet & HFTNN & HFCNN & TCNN \\
		\midrule
		 MNIST & 23.00\%$\pm$2.91\% & 23.00\%$\pm$2.91\% &23.00\%$\pm$2.91\% & 93.06\%$\pm$2.45\% &25.00\%$\pm$2.90\% & 94.37\%$\pm$1.21\% & \textbf{98.86\%$\pm$1.02\%} \\
		FASHIONMNIST & 25.00\%$\pm$2.90\% & 25.00\%$\pm$2.90\% &25.00\%$\pm$2.90\% &86.90\%$\pm$2.50\% &25.00\%$\pm$2.90\% & 91.22\%$\pm$1.76\% & \textbf{94.09\%$\pm$1.39\%} \\
		CIFAR10 & 25.00\%$\pm$2.90\% & 25.00\%$\pm$2.90\% & 25.00\%$\pm$2.90\% & 67.89\%$\pm$2.74\% & 25.00\%$\pm$2.90\% & 77.04\%$\pm$1.68\% & \textbf{79.54\%$\pm$1.38\%} \\
        CIFAR100 & 25.00\%$\pm$2.90\% & 25.00\%$\pm$2.90\% & 25.00\%$\pm$2.90\% & 72.10\%$\pm$2.84\% & 25.00\%$\pm$2.90\% & 73.39\%$\pm$1.87\% & \textbf{76.44\%$\pm$1.71\%} \\
        ImageNet & 25.00\%$\pm$2.90\% & 25.00\%$\pm$2.90\% & 25.00\%$\pm$2.90\% & 78.44\%$\pm$2.35\% & 25.00\%$\pm$2.90\% & 76.09\%$\pm$2.01\% & \textbf{81.02\%$\pm$1.78\%} \\
        CelebA & 25.00\%$\pm$2.90\% & 25.00\%$\pm$2.90\% & 25.00\%$\pm$2.90\% & 87.05\%$\pm$2.24\% & 25.00\%$\pm$2.90\% & 88.66\%$\pm$1.98\% & \textbf{90.97\%$\pm$1.56\%} \\
		\bottomrule
	\end{tabular}}
\end{table*}
\begin{table*}[!t]
	\centering
	\renewcommand{\arraystretch}{0.6}
	\caption{Loss of the seven models to random normal distribution noise attacks}
	\centering
	\scalebox{0.8}{
    \setlength{\tabcolsep}{7mm}
	\begin{tabular}{c c c c c c c c}
		\toprule
		Dataset & MLP & QiMLP & FATTNN& HQNet & HFTNN & HFCNN & TCNN \\
		\midrule
		MNIST& nan &nan & nan &0.0086 & nan & 0.0035 & \textbf{0.0024} \\
		FASHIONMNIST& nan & nan &nan&0.0056&nan& 0.0049 & \textbf{0.0046} \\
		CIFAR10& nan & nan &nan&0.084&nan& 0.073 & \textbf{0.046} \\
        CIFAR100&nan & nan &nan&0.076&nan& 0.056&\textbf{0.049} \\
        ImageNet& nan & nan &nan&0.123&nan& 0.097 & \textbf{0.078} \\
        CelebA& nan & nan &nan&0.085&nan& 0.053&\textbf{0.045} \\
		\bottomrule
	\end{tabular}}
\end{table*}
\begin{table*}[!t]
	\centering
	\renewcommand{\arraystretch}{0.6}
	\caption{Gradients of the loss to a certain attacked parameter of the compared models on CIFAR datasets}
	\centering
	\scalebox{0.8}{
    \setlength{\tabcolsep}{7mm}
	\begin{tabular}{c c c c c c c}
		\toprule
		noise distribution attack& MLP &QiMLP&FATTNN& HFTNN & Resnet & Densenet\\
		\midrule
		 random normal distribution & nan &nan &nan & nan & nan & nan \\
		 random uniform distribution & nan &nan &nan & nan & nan & nan \\
		\bottomrule
	\end{tabular}}
\end{table*}
\begin{table*}[!t]
	\centering
	\renewcommand{\arraystretch}{0.6}
	\caption{Accuracy of the four models. * means enhanced datasets}
	\label{table_example}
	\centering
	\scalebox{0.8}{
    \setlength{\tabcolsep}{8mm}
	\begin{tabular}{c c c c c}
		\toprule
		Dataset  & TCNN & TCNN2 & TCNN3 & TCNN4\\
		\midrule
		MNIST & \textbf{98.95\%$\pm$1.02\%} & 96.35\%$\pm$1.34\% & 96.42\%$\pm$1.36\% & 95.38\%$\pm$1.46\%\\
		FASHIONMNIST  &\textbf{93.90\%$\pm$1.64\%} & 91.43\%$\pm$2.12\% & 90.78\%$\pm$2.42\% &90.67\%$\pm$2.45\%\\
		MNIST*   & \textbf{98.98\%$\pm$1.20\%} & 96.79\%$\pm$1.25\% & 96.56\%$\pm$1.23\% & 96.10\%$\pm$1.67\%\\
        FASHIONMNIST* & \textbf{93.79\%$\pm$1.71\%} & 91.09\%$\pm$2.23\% & 91.21\%$\pm$1.98\% & 91.45\%$\pm$2.06\%\\
        CelebA & \textbf{92.68\%$\pm$0.93\%} &88.64\%$\pm$0.87\%&89.44\%$\pm$0.82\%&89.65\%$\pm$0.81\% \\
        CelebA* & \textbf{92.96\%$\pm$0.87\%} &90.03\%$\pm$0.96\%&90.15\%$\pm$0.95\%&89.34\%$\pm$0.99\% \\
		\bottomrule
	\end{tabular}}
\end{table*}
\begin{figure}[!h]
	\centering
    \includegraphics[width=3.3in]{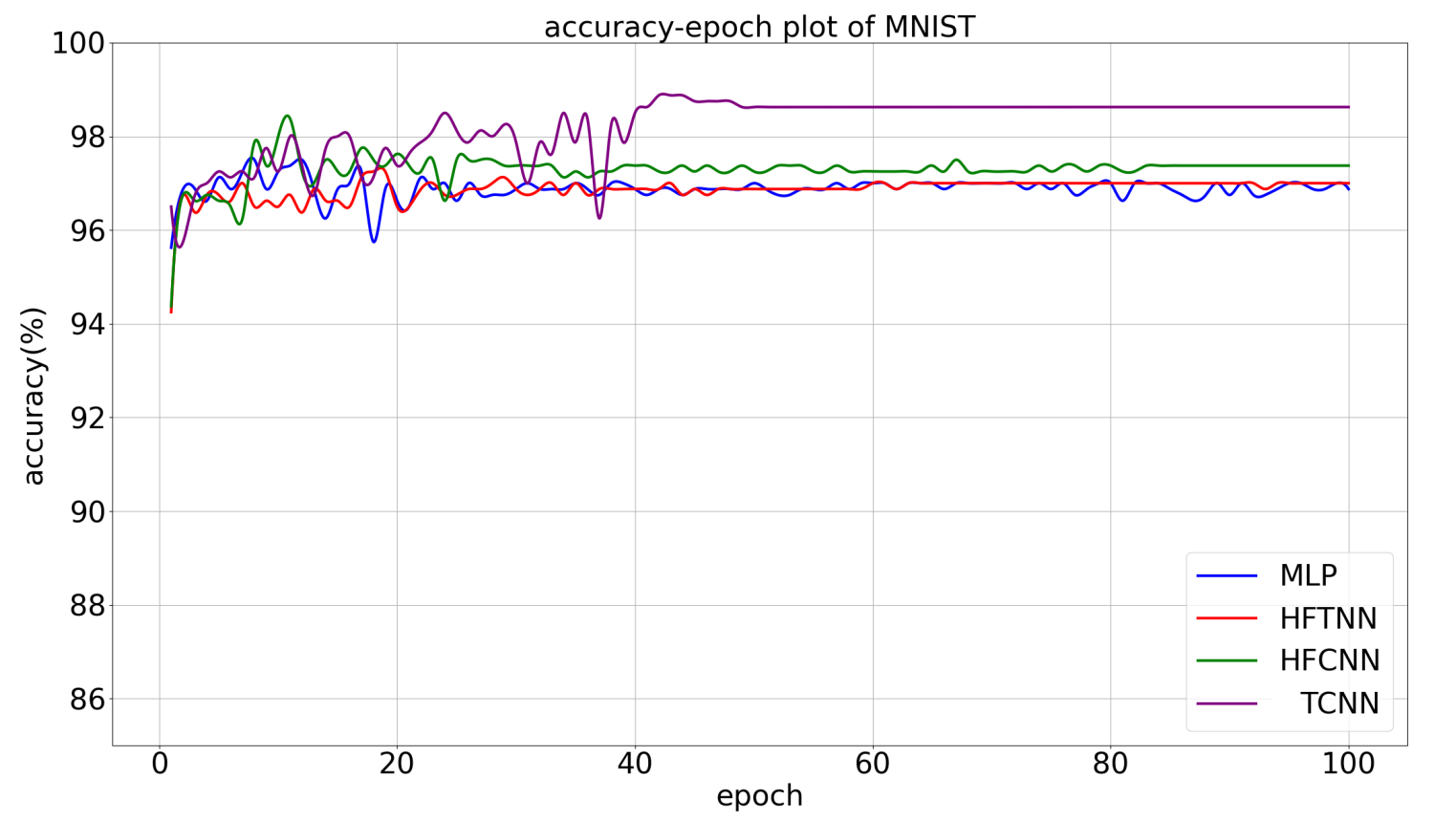}
	\caption{Accuracy curves of the four models on clean MNIST data. All the curves fluctuate first, and then reach convergence. The accuracy of TCNN is the highest.}
\end{figure}
\begin{figure}[!h]
	\centering
    \includegraphics[width=2.8in]{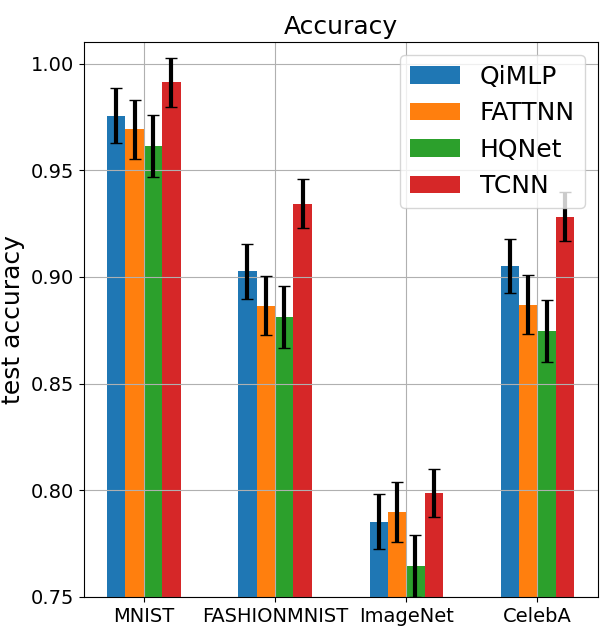}
	\caption{Accuracy charts of the four models on assorted datasets. The accuracy of TCNN is the highest.}
\end{figure}
 We implement different groups of experiments with clean and enhanced data. Fig.5 demonstrates the accuracy curves of the four models on MNIST. Table 2 provides the average accuracy, loss values and CPU running time of TCNN and all the compared models. From the tables, we can see that TCNN showcases a little more advantages than other models on accuracy $\mu$ and loss. We also calculate the standard deviation of accuracy $E_{\mu}$ with the formula \cite{2}:
\begin{equation}
    E_{\mu} = Z_{score}\times \sqrt{\frac{\mu \cdot(1-\mu)}{Q}}, 
\end{equation}
where $Q$ represents the number of samples in the testing dataset. Considering the 95\% confidential intervals, $Z_{score} = 1.96$ \cite{2}. 
What's more, the loss values of TCNN are almost the lowest among the compared models. In comparison with HFTNN, the paralleled circuit models in TCNN facilitate generalization capability of it. In contrast with the MLP and HFCNN, TCNN possess lower parameter complexity. Simultaneously, the time complexity of TCNN is comparable to the state-of-the-art MLP and FATTNN, which is lower than that of HFCNN and HQNet. More outcomes in comparison with Resnet and Densenet are in the supplementary material that also demonstrate the advantages of TCNN.\\ 
\textbf{Robustness test.} 
Fig.6 demonstrates the accuracy of the four models on datasets with random normal distribution noise. Other results in the supplementary material demonstrate the accuracy, loss values and CPU running time of the compared models. They demonstrate that the learning ability of TCNN is marginally stronger than the other three models according to accuracy and loss values.\\
\textbf{Sensitive analysis.} 
To evaluate the stability and sensitivity of our TCNN, we implement experiments with different enhanced or noisy datasets to obtain multiple average accuracy values to calculate the variance of them after the model converge to evaluate the stability of the model. The results are shown in the table 6, which showcases the high stability and low sensitivity of our TCNN. \\
\textbf{Case study.}
In addition, as the residual circuit model in the previous work demonstrate unique robustness to adversarial noise attack \cite{46}, we also conduct experiments to validate its robustness to random noise parameter attacks. Table 4 and 5 separately showcase the average accuracy values and the loss values of the four models. They indicate that the accuracy of the MLP, HFTNN, FATTNN, QiMLP and the two convolutional networks remains about 20\%, while the loss values of them rises a lot, finally reaching nan in the program codes, which implies that they are not capable of resisting noise attacks. Since HQNet and HFCNN contains circuit models that could resist noise attacking, they are able to learn features but with lower accuracy than TCNN. Table 6 demonstrates the infinite gradients of the loss function to the attacked parameters of the compared models, which indicates that gradient explosion results in the failure of the model testing. There are more advantageous results about robustness of our TCNN in comparison with the two convolutional networks in the supplementary material.\\
\textbf{Ablation study.}
As we devise paralleled circuit models with activations and information fusion layer to enhance its performance, we conduct ablation study comparing TCNN and another three hybrid tensor circuit networks (TCNN2, TCNN3 and TCNN4) whose architectures are almost the same as TCNN. TCNN2 possess paralleled circuit framework and information fusion layer but no activation. TCNN3 possess information fusion layer but it only has serial activated circuit framework. TCNN4 possess activation and paralleled circuit architectures but no information fusion layer. The accuracy values on different datasets are shown in table 7. It indicates the importance of the parallelism and activation in the circuit model and the information fusion layer.

\section{Conclusion and discussion}
In this paper, we propose a tensor circuit neural network (TCNN) consisting of a tensor neural network with MPO factorization method and paralleled residual circuit models with unitary evolutions and activations. By analyzing its framework and parameter complexity, one advantage of TCNN lies in its low parameter complexity. What's more, groups of experiments demonstrate that TCNN possess more superiority than the MLP, QiMLP, FATTNN, HQNet, HFTNN, HFCNN and two convolutional networks on image classification tasks with clean, enhanced and noisy datasets. In the meantime, the time complexity of TCNN is lower than that of HFCNN and comparable to the MLP and FATTNN. More significantly, groups of case study demonstrate that our TCNN surpasses other compared models a lot in resistance to random noise parameter attacks due to its distinguished ability to prevent gradient explosion. Meanwhile, our TCNN shows higher stability and lower sensitivity than the compared models. Finally, through an ablation study, we evaluate the advantage of the parallelism, activation operation in the circuit models and the information fusion layer in TCNN. On the basis of these comprehensive strengths, it makes sense to assess TCNN on complicated industrial scenarios with the interference of assorted noises and deploy it on practical electrical hardware to accelerate its industrialization \cite{59,60}. 



\bibliography{ref-AAAI}
\end{document}